# AUTOMATIC SPEECH RECOGNITION FOR THE IKA LANGUAGE

**DR. UCHENNA NZENWATA, OGBUIGWE, DANIEL UDOKA**

Department of Computer Science,
School of Computing and Engineering Sciences

## ABSTRACT

We present a cost-effective approach for developing Automatic Speech Recognition (ASR) models for low-resource languages like Ika. We fine-tune the pretrained wav2vec 2.0 Massively Multilingual Speech Models on a high-quality speech dataset compiled from New Testament Bible translations in Ika. Our results show that fine-tuning multilingual pretrained models achieves a Word Error Rate (WER) of 0.5377 and Character Error Rate (CER) of 0.2651 with just over 1 hour of training data. The larger 1 billion parameter model outperforms the smaller 300 million parameter model due to its greater complexity and ability to store richer speech representations. However, we observe overfitting to the small training dataset, reducing generalizability. Our findings demonstrate the potential of leveraging multilingual pretrained models for low-resource languages. Future work should focus on expanding the dataset and exploring techniques to mitigate overfitting.

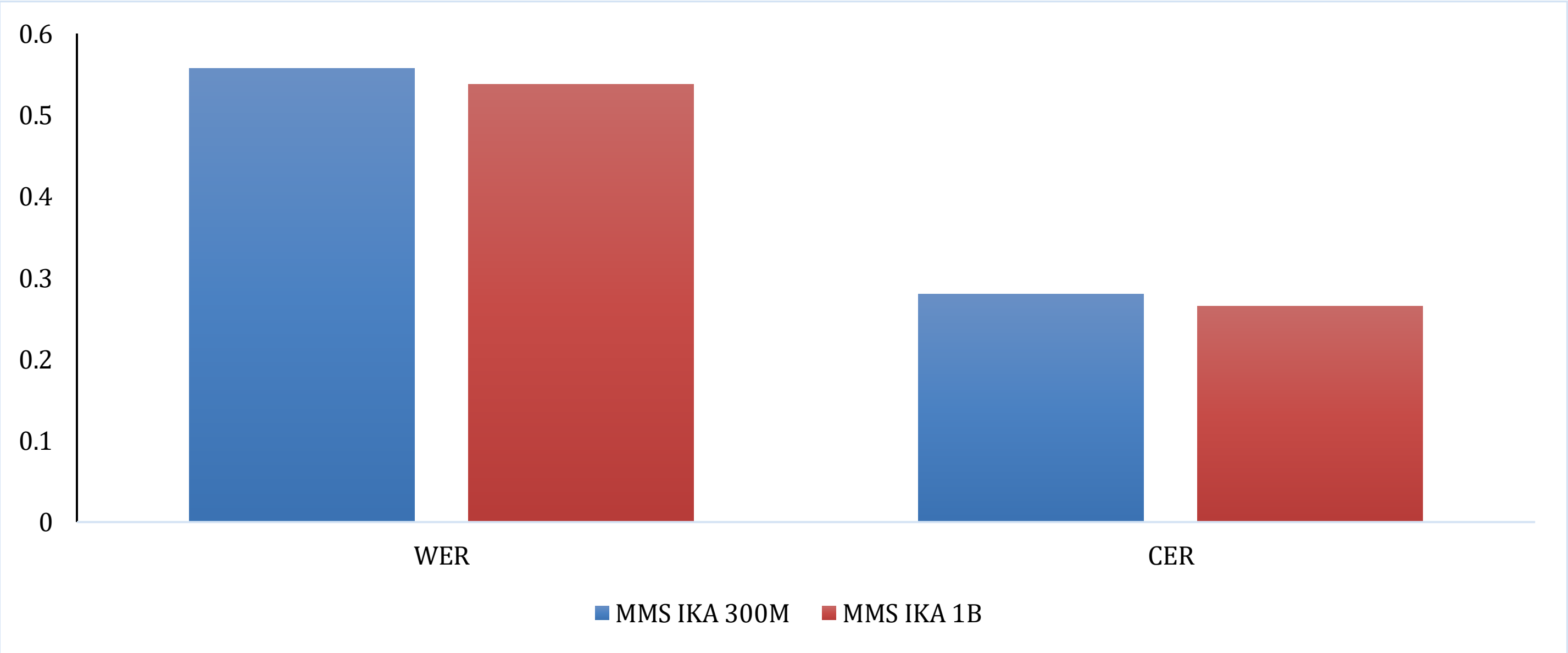

0.1. WER and CER performance comparison of the 965 million (MMS-1B) and 317 million (MMS-300M) models on our validation set. Lower values are better.

# I. INTRODUCTION

## Background

Automatic Speech Recognition (ASR) systems data back to the 1950s [1], [2]. These systems have become integral in enabling machines to understand and process human language. They facilitate applications from voice-activated assistants to transcription services. Recent advancements in ASR technology, particularly with models like wav2vec 2.0, have significantly improved performance by leveraging unsupervised pre-training on large-scale unlabeled audio data followed by fine-tuning on labeled datasets [3]. The wave2vec 2.0 model, developed by Meta AI, has demonstrated superior performance across various languages by learning robust audio representations that can be fine-tuned for specific ASR tasks.

## Significance

Despite these advancements, ASR development for low-resource languages remains a significant challenge. Many languages lack the extensive labeled datasets required for effective training of ASR systems [4]. The Ika language, spoken by a minority in Nigeria, is one such 'low resource' languages. Although Ika is a stable language that is still taught to and spoken by children in Agbor and other Ika communities, data scarcity has hindered the development of sophisticated encoding tools and language technology [5], [6]. The development of an ASR system for Ika is crucial for preserving and promoting linguistic diversity, providing speakers with modern technological tools that support their language, and facilitating the documentation and study of Ika

## Objective

This study aims to fine-tune a pre-trained wav2vec 2.0 model on a newly created dataset for the Ika language, derived from audio recordings of religious texts. Specifically, we utilize readings of the chapters of John and Mark from the Ika Bible, segmented into verse-level audio clips using the MMS force aligner. By aligning these audio clips with their corresponding transcripts, we create a high-quality dataset for fine-tuning the wav2vec 2.0 model. Our goal is to develop an ASR system that accurately recognizes and transcribes spoken Ika, demonstrating the potential of leveraging pre-trained models and innovative data alignment techniques for low-resource languages.

# II. RELATED WORK

## Multilingual ASR Models

The development of multilingual ASR systems has seen significant progress. One notable project is Meta's Massively Multilingual Speech (MMS) initiative, which has scaled speech technology across over 1,100 languages. This project includes sophisticated data alignment processes and fine-tuning on large multilingual datasets, like the approach taken in this study with the Ika Bible audio readings [7]. The MMS project highlights the importance of multilingual data and robust alignment techniques in improving ASR performance for low-resource languages.

## Wav2vec 2.0

Wav2vec 2.0, developed by Meta AI, is a leading model in speech recognition. It leverages self-supervised learning to pre-train on vast amounts of unlabeled audio data, followed by fine-tuning on labeled datasets. Research shows that wave2vec 2.0 achieves state-of-the-art results on several benchmarks, making it a strong candidate for low-resource language ASR [3]. Studies demonstrate its effectiveness in low-resource environments, such as deployment on edge devices like the Raspberry Pi, highlighting its robustness and adaptability [8].

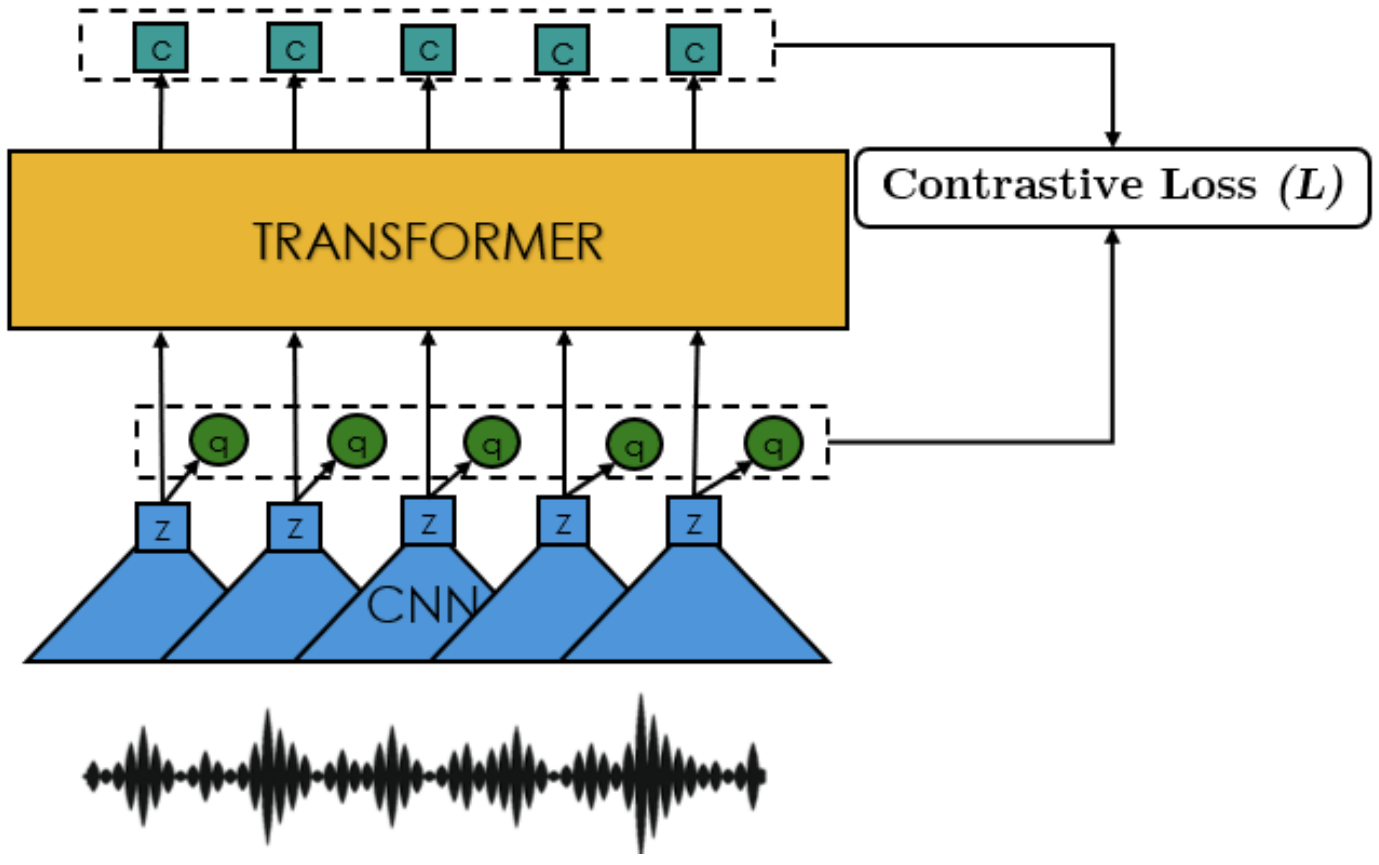

2.1. Self-Supervised pretraining phase of wav2vec 2.0 models.

## End-to-End ASR Systems

OpenAI's Whisper model represents a significant advancement in end-to-end ASR systems [9]. Unlike modular approaches, Whisper integrates the entire ASR pipeline into a single neural network, simplifying the training and deployment process. Comparisons between Whisper and modular approaches like wav2vec 2.0 highlight the efficiency and simplicity of end-to-end models, especially in handling diverse accents and languages. However, modular approaches offer greater flexibility for integrating additional language models and fine-tuning for specific tasks. For example, wav2vec2 models often outperform Whisper when fine-tuned for specific needs, such as recognizing child speech. While end-to-end models like Whisper streamline many aspects of ASR, they are data-hungry, and adapting them to languages with limited datasets remains an area requiring further exploration [10].

## Sequence Modelling with Connectionist Temporal Classification

Connectionist Temporal Classification (CTC) serves as a pivotal algorithm for sequence-to-sequence problems, notably in automatic speech recognition (ASR) and handwriting recognition. Its core functionality allows neural networks to learn alignments between input sequences (such as audio signals) and their corresponding output sequences (text), despite differences in sequence lengths. CTC uniquely incorporates a "blank" token within its framework, facilitating the model's ability to predict character repetitions and omissions in the output sequence, thereby enhancing its predictive accuracy and adaptability.

Pretrained models like the MMS models can already map speech features (e.g., spectrograms) to a sequence of probabilities over possible output units (phonemes or letters). During fine-tuning, we introduce CTC loss into the model. The CTC objective for a single (X,Y) pair is given by:

$$p(Y \mid X) = \sum_{A \in A_{X,Y}} \prod_{t=1}^{T} p_t(a_t \mid X)$$

where $A(X,Y)$ represents the set of all possible alignments A between the input X and the output Y, T is the length of the input sequence, and $pt\ (at \mid X)$ is the probability of the alignment a_t at time step t given the input $X$.

To compute the CTC loss $L_{CTC}$, we take the negative log probability:

$$L_{CTC} = -\log p(Y \mid X)$$

CTC analyses the alignment between the actual transcript (text) and the model's predicted probability sequence. It calculates the cost (loss) associated with this misalignment, aiming to minimize this loss by adjusting the model's weights through the Adam optimizer [11] during training. CTC excels because it can handle speech segments of varying lengths corresponding to different word durations without needing pre-segmentation. It tolerates potential errors in the model's predictions, such as inserting or deleting phonemes in the output sequence, which is helpful for dealing with imperfections in speech recognition [12], [13].

## Cross-Lingual Representation Learning

Cross-lingual speech representation learning has been a key area of focus. Conneau *et al* [14] introduced XLSR (Cross-Lingual Speech Representation Learning), a model designed to pretrain on raw speech waveforms from multiple languages. This model significantly outperforms monolingual models by leveraging shared latent discrete speech representations across languages. On benchmarks such as CommonVoice and BABEL, XLSR reduces the phoneme error rate by 72% and improves the word error rate by 16% [15].

**Low-Resource Language ASR**

Seo *et al* [16] developed an ASR model for the Manchu language using Wav2Vec2-XLSR-53, addressing data scarcity through data augmentation techniques like adding background noise, reverberation, clipping audio, and removing audio segments. This approach resulted in significant reductions in Character Error Rate (CER) and Word Error Rate (WER). Their approach demonstrates the potential effectiveness for extremely low-resource languages like Ika [16].

**Adapter-Based Tuning for Pretrained Models**

He *et al* (2021) examined adapter-based tuning as a parameter-efficient alternative to fine-tuning pretrained language models. This involves adding lightweight adapter modules, which are fine-tuned for specific tasks. This method is effective for low-resource and cross-lingual tasks and mitigates forgetting issues, aligning well with our research objectives for efficiently adapting pretrained models for Ika without extensive computational resources [17].

**Adapting Multilingual Models for Under-Resourced Languages**

Nowakowski *et al* (2021) explored adapting multilingual speech representation models for new, under-resourced languages using the example of Sakhalin Ainu. They showed that continued pretraining on target language data, combined with leveraging data from related languages, significantly improves performance. For Ika, this suggests that similar strategies could enhance speech recognition accuracy. However, due to the limited data, we will not employ continued pretraining in our strategy [18].

**Summary**

From these studies, several strategies and insights emerge that are applicable to our goal of advancing speech recognition for the Ika language:

1. Cross-Lingual Transfer: Utilizing cross-lingual pretraining methods, as demonstrated by XLSR, could significantly enhance Ika speech recognition by leveraging shared representations across multiple languages.

2. Low-Resource Strategies: Employing data augmentation techniques similar to those used for Manchu ASR can help mitigate the challenges posed by limited Ika language data.

3. Parameter Efficiency: Utilizing adapter-based tuning can efficiently adapt pretrained models for Ika, making the most of limited computational and data resources.

4. Continued Pretraining: Although not employed in our strategy due to limited data, continued pretraining on Ika-specific data and incorporating data from related languages could enhance speech recognition system accuracy.

By integrating these strategies, the development of a robust and effective speech recognition system for the Ika language can be significantly advanced.

## III. METHODOLOGY

**Dataset**

The initiation of our research endeavor was hindered by the absence of publicly available datasets for the Ika language, necessitating the creation of a bespoke dataset. In the realm of speech recognition, the gold standard for dataset creation involves collaboration with language experts and native speakers to record a diverse array of speech samples. However, due to resource constraints, we opted for an alternative approach, leveraging open online sources to construct our dataset.

Recognizing the significant efforts of religious organizations in bridging language barriers, we selected religious texts as the foundation for our dataset. The Ika Bible, provided by the Mission for Language Translation (MLT), offered a valuable resource for our research. The dataset comprises comprehensive audio recordings of the New Testament, featuring verse-by-verse readings of all chapters. This selection ensured a substantial and diverse dataset, conducive to effective speech recognition model training.

The raw audio recordings were sourced from the Ika Bible site, while text transcriptions were obtained from (link unavailable) The preprocessing phase, executed on T4 GPUs via Google Colab, involved the normalization of text files through the removal of punctuation and numerical digits, as well as the elimination of duplicate or empty lines. Alignment and segmentation were performed utilizing scripts from the MMS data prep repository on fairseq's GitHub, culminating in the generation of a manifest JSON file via the MMS Force Aligner (MMS_FA).

**Collecting and Organizing Raw Data**

The raw dataset comprises three files: `IKKTBLN1DA`, `IKKTBLN2DA`, and `ikkNT_readaloud`. The audio files are stored in folders named `IKKTBLN-` and are available in MP3 format. These are categorized as dramatic (containing music and sound effects) and non-dramatic (plain audio). For this task, only the non-dramatic audio files were used to simplify preprocessing.

The text files in the `ikkNT_readaloud` folder contain an introduction and the verses of each chapter. The dataset includes 260 chapters, each with an audio reading and text transcript. A Python script was used to regularize the file names for consistency, making batch processing more straightforward. For example, "`B01__01_Matthew___IKKTBLN1DA.MP3`" and "`ikkNT_070_MAT_01_read.txt`" were renamed to "`Matthew_01.mp3`" and "`Matthew_01.txt`" respectively.

**Generating Audio and Text Pairs**

Only the books of Mark (16 chapters) and John (21 chapters) were selected for fine-tuning, reducing the dataset to 37 chapters. This choice was made to avoid overfitting the pretrained models on the entire New Testament dataset.

**Audio Processing: Pre-Segmentation**

Long audio files (average of 4 minutes) were segmented to align with the verse-level units of the text. This segmentation was necessary due to GPU memory constraints and the computational complexity of processing long sequences in transformer models.

Steps for Pre-Segmentation: All audio files were normalized using Audacity to adjust the amplitude to a standard level, ensuring consistent volume across recordings. Audio segments were then converted to single-channel (mono) WAV files with a 16 kHz sampling rate, compatible with the wav2vec 2.0 model architecture.

Automated segmentation was implemented using the Multilingual Massively Multitask Speech (MMS) Data Preparation Library and the NVIDIA NeMo Toolkit. This method was chosen for its effectiveness in handling diverse languages and large-scale religious texts.

**Force Alignment Process**

To facilitate the force alignment process, we utilized Google Colab's T4 GPUs to install necessary libraries, including torchaudio and sox. We then cloned the Fairseq and uroman GitHub repositories, enabling us to load the MMS Force Alignment model. To accommodate custom argument passing, we modified key scripts, namely align and segment script, the normalization script and paths within the notebook. Subsequently, we executed the alignment process for all audio and text pairs corresponding to the books of John and Mark. This process yielded a manifest JSON file containing essential metadata, as well as segmented audio files.

**Filtering: Removing Low-Quality Segments**

The manifest JSON file contains metadata for each audio segment, including:
1. `audio_start_sec`: Start time in seconds.
2. `audio_filepath`: Path to the audio segment.
3. `duration`: Duration in seconds.
4. `text`: Transcribed text.
5. `normalized_text`: Normalized transcribed text.
6. `uroman_tokens`: Romanized tokens from the text.

In conjunction with NVIDIA NeMo Speech Data Explorer, we leveraged this metadata to assess segment quality, evaluating metrics such as word count, character count, word rate, and character rate. Visual representations, including spectrogram and waveform displays, further aided in segment evaluation. We then conducted manual listening and quality tagging, categorizing segments as "High," "Low," or "Fixable." Only "High" quality segments were selected for fine-tuning purposes. The dataset was subsequently divided into training and testing sets in an 80/20 ratio, accompanied by the creation of a metadata CSV file containing file path, transcription, split designation (train/test), and file name.

**Model Training**

The pre-trained MMS-1B and MMS-300M models were employed as the foundation for our Ika speech recognition system, utilizing their default configurations as a starting point. To fine-tune and evaluate these models, we used Python libraries, including datasets, transformers, torchaudio, huggingface, and jiwer. The performance of the models was assessed using two key metrics: Word Error Rate (WER) and Character Error Rate (CER).

**Training Parameters**

Our review of similar research showed us the critical training parameters that needed calibration [19], [20], [21]. The batch size was set to 4, to balance between memory usage and the training time. The models were trained between 10 and 30 epochs, aiming to achieve a balance between learning from the limited dataset and avoiding overfitting. The learning rate, a crucial parameter controlling the step size of the optimizer, was set to 3e-4 and adjusted using a warmup strategy over 100 steps to stabilize the early stages of training.

**Data Collator**

To accommodate the unique requirements of speech input and output modalities, a custom data collator script was written. This collator dynamically padded inputs and labels, treating input values and labels differently to ensure effective processing and alignment. By using this custom data collator, we ensured that the models received well-formatted and consistent input data, enabling them to learn the nuances of the Ika language effectively.

**Evaluation Metric**

Word Error Rate (WER), Character Error rate (CER), Training and Validation loss are the primary evaluation metrics measured during training. calculated by decoding predictions and comparing them with the reference transcriptions.

**Model Configuration**

The pretrained `mms-1b` and `mms-300m` checkpoints were loaded with configurations for dropout, SpecAugment's masking dropout rate, layer dropout, and learning rate, ensuring stable training. Gradient checkpointing and loss reduction to "mean" were enabled to save GPU memory.

## IV. EXPERIMENT

We fine-tuned both sizes of the MMS Wav2Vec 2.0 model, one with 1 billion parameters `("facebook/mms-1b")` and the other with 300 million parameters `("facebook/mms-300m")`, to develop the speech recognition system for the Ika language. Both models were fine-tuned on the same dataset using a consistent set of training parameters to ensure comparability of results. The training parameters included a dropout rate of 0.0, a mask time probability of 0.05, and no layer dropout, maintaining all neural network layers' contributions during training. The CTC loss reduction was set to "mean" to average the loss over the batch for stable gradient updates, and the vocabulary size was matched to the tokenizer length, ensuring all potential Ika phonemes were representable.

The batch size was initially set to 4, with gradient accumulation starting at 2 and later increased to 4 in subsequent iterations to improve stability. The learning rate was fixed at 3e-4 with a warmup period of 500 steps to gradually reach the target learning rate, reducing the risk of early training instability. The models were trained on a single NVIDIA L4 GPU, equipped with 22GB of RAM.

## V. RESULTS

### 1 billion Parameter Model (MMS-1B)

First Iteration: The initial training iteration for the 1 billion parameter model was set for 30 epochs and took approximately 1 hour, 1 minute, and 55 seconds to complete. This iteration exhibited significant overfitting, as evidenced by the training loss approaching zero while the validation loss continued to rise. Despite this, the model achieved a Word Error Rate (WER) of 0.477553 and a Character Error Rate (CER) of 0.241756.

Second Iteration: To address the overfitting observed in the first iteration, the number of epochs was reduced to 20. This iteration lasted 34 minutes and 49 seconds. Although this adjustment helped reduce overfitting, the WER and CER metrics slightly worsened, with the WER increasing to 0.556981 and the CER to 0.280248.

Third Iteration: Further reducing the number of epochs to 13 resulted in a more stable training process, minimizing overfitting. The final performance metrics for this iteration were a WER of 0.537741 and a CER of 0.265053, indicating improved stability.

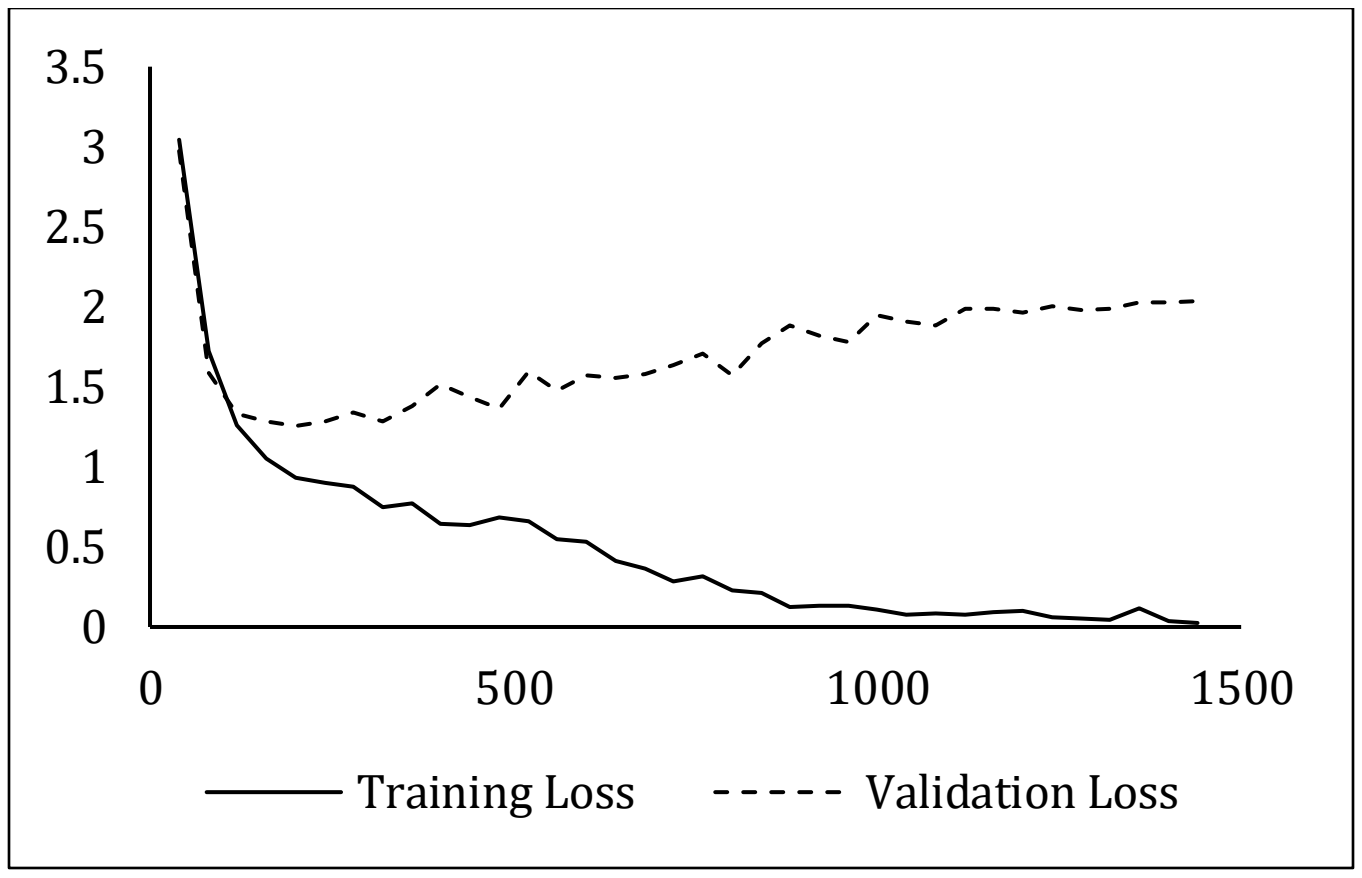

Figure 5.1. Training and validation loss for the 1 billion parameter model over multiple iterations.

### 300 million Parameter Model (MMS-300m)

First Iteration: The initial fine-tuning iteration for the 300 million parameter model was set for 20 epochs and lasted 17 minutes and 28 seconds. This model showed signs of underfitting, with both the WER and CER remaining high throughout training, each at 1.000000.

Second Iteration: Adjustments were made to the training parameters in the second iteration, including setting all dropout rates to 0.0 and the mask time probability to 0.0. These adjustments led to improvements in performance, with the WER decreasing to 0.617662 and the CER to 0.287001. However, the model required more training steps to achieve these results.

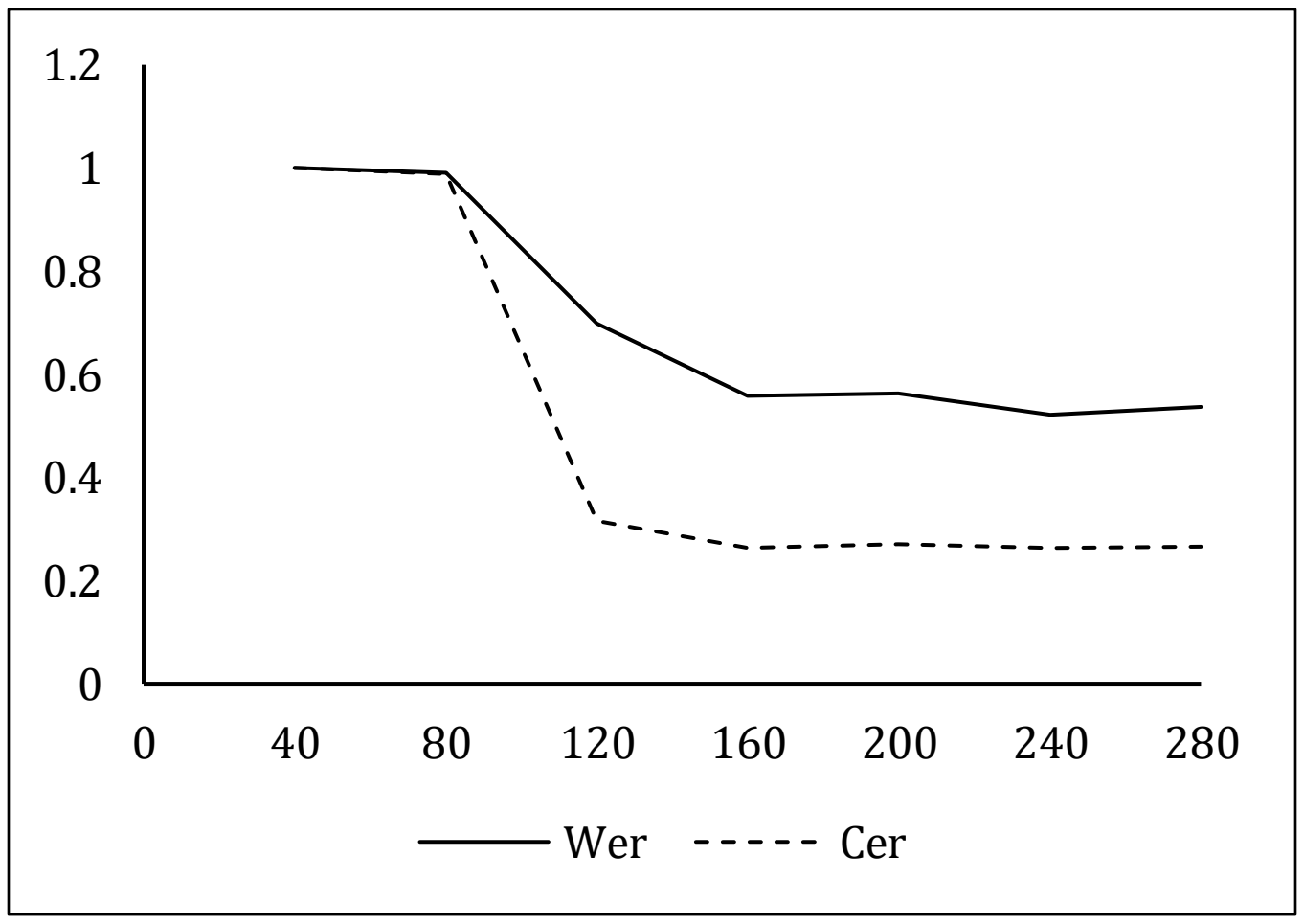

Figure 4.8: Training and validation loss for the 300 million parameter model.

### Comparison

The comparison between the two models showed that the 1 billion parameter model outperformed the 300 million parameter model, achieving better WER and CER metrics in fewer training steps and epochs. This highlights the trade-off between model complexity and performance, particularly in low-resource settings.

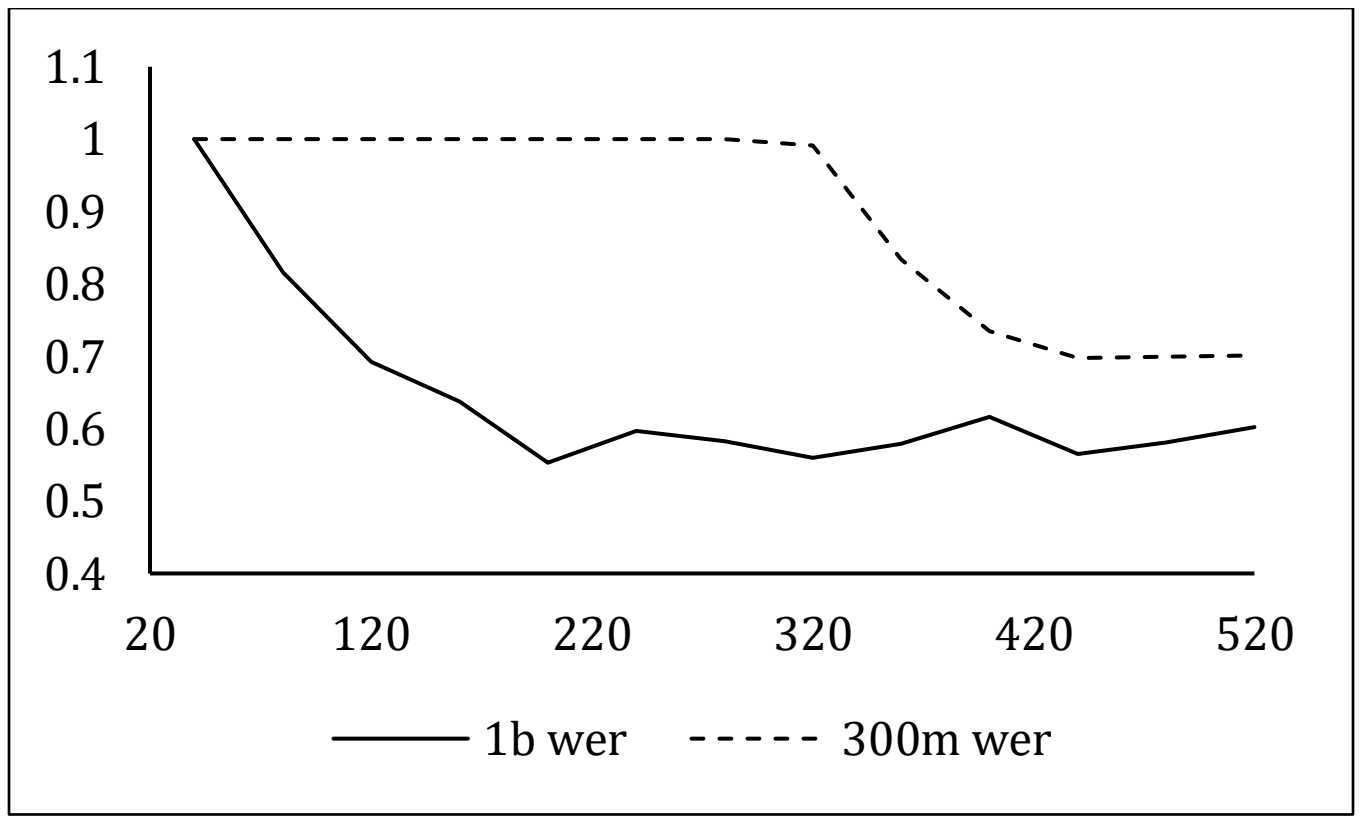

Figure 5.3. Comparison of WER and CER across both models.

## VI. DISCUSSION

The results highlight the effectiveness of fine-tuning multilingual pretrained models for low-resource languages like Ika. The larger 1 billion parameter model demonstrated better performance and required fewer training steps to achieve optimal WER and CER compared to the 300 million parameter model. This suggests that richer pretrained representations significantly benefit low-resource ASR tasks.

### Challenges

1. Overfitting: The initial runs with the 1 billion parameter model showed significant overfitting, which was mitigated by reducing the number of epochs.

2. Underfitting: The 300 million parameter model initially displayed underfitting, necessitating adjustments in training parameters.

3. GPU Memory Constraints: Frequent "GPU Out of Memory" errors were encountered, particularly with the 1b model. Switching to a larger L4 GPU helped alleviate this issue, but maintaining smaller per-device training batch sizes was necessary.

### Future Work

1. Dataset Expansion: Increasing the size of the Ika language dataset would likely improve model performance and reduce overfitting.

2. Data Augmentation: Implementing data augmentation techniques could enhance the model's robustness and generalization capabilities.

3. Semi-Supervised Learning: Exploring semi-supervised learning approaches could leverage unlabeled data to further improve ASR performance in low-resource settings.

4. Model Scalability: Investigating the scalability of these models in terms of both performance and computational efficiency would be valuable for broader applications in low-resource languages.

In conclusion, while the fine-tuned models demonstrated promising results for Ika language ASR, expanding the dataset and exploring advanced training techniques are essential next steps to enhance model performance and generalization.

## VII. CLOSING

### Summary

This study explored the fine-tuning of multilingual pretrained MMS Wav2Vec 2.0 models with 1 billion and 300 million parameters for automatic speech recognition (ASR) in the Ika language. Through iterative adjustments in training parameters, we aimed to optimize the performance of these models on a small Ika language dataset. The 1 billion parameter model demonstrated superior performance, achieving lower Word Error Rate (WER) and Character Error Rate (CER) in fewer training steps compared to the 300 million parameter model. Key challenges included overfitting in the larger model and also reduced underfitting in the smaller model. Training parameters were carefully tuned to avoid those challenges.

## Impact

The findings of this study highlight the potential of leveraging large multilingual pretrained models for ASR in low-resource languages. The richer pretrained representations in larger models significantly enhance their ability to adapt to new languages with limited training data. This work underscores the importance of model scalability and efficient training practices in developing robust ASR systems for low-resource languages, which are often underrepresented in the field of speech recognition.

## Future Directions

Future research should focus on expanding the Ika language dataset to improve model performance and reduce overfitting. Implementing data augmentation techniques and exploring semi-supervised learning approaches could further enhance the robustness and generalization capabilities of the ASR models. Additionally, investigating the scalability of these models in terms of both performance and computational efficiency will be crucial for broader applications in low-resource languages. Finally, developing dedicated test sets for more accurate evaluation of model performance in real-world scenarios is essential for advancing ASR technology in low-resource settings.

## VIII. REFERENCE

[1] K. Garrett, "The History of Speech Recognition," Transcribe. Accessed: Jul. 08, 2024. [Online]. Available: https://transcribe.com/blog/the-history-of-speech-recognition

[2] V. Rotter, "Exploring the Evolution of Speech Recognition: From Audrey to Alexa," Audeering.

[3] A. Baevski, H. Zhou, A. Mohamed, and M. Auli, "wav2vec 2.0: A Framework for Self-Supervised Learning of Speech Representations," Oct. 22, 2020, *arXiv*: arXiv:2006.11477. Accessed: Mar. 04, 2024. [Online]. Available: http://arxiv.org/abs/2006.11477

[4] A. Yeroyan and N. Karpov, "Enabling ASR for Low-Resource Languages: A Comprehensive Dataset Creation Approach," Jun. 03, 2024, *arXiv*: arXiv:2406.01446. Accessed: Aug. 07, 2024. [Online]. Available: http://arxiv.org/abs/2406.01446

[5] "Ika | Ethnologue Free," Ethnologue (Free All). Accessed: Mar. 18, 2024. [Online]. Available: https://www.ethnologue.com/language/ikk/

[6] "How many languages are endangered? | Ethnologue Free." Accessed: Mar. 18, 2024. [Online]. Available: https://www.ethnologue.com/insights/how-many-languages-endangered/

[7] V. Pratap *et al.*, "Scaling Speech Technology to 1,000+ Languages," May 22, 2023, *arXiv*: arXiv:2305.13516. Accessed: Sep. 10, 2023. [Online]. Available: http://arxiv.org/abs/2305.13516

[8] S. Gondi, "Wav2Vec2.0 on the Edge: Performance Evaluation," 2022, *arXiv*. doi: 10.48550/ARXIV.2202.05993.

[9] A. Radford, J. W. Kim, T. Xu, G. Brockman, C. McLeavey, and I. Sutskever, "Robust Speech Recognition via Large-Scale Weak Supervision," Dec. 06, 2022, *arXiv*: arXiv:2212.04356. Accessed: Mar. 02, 2024. [Online]. Available: http://arxiv.org/abs/2212.04356

[10] R. Jain, A. Barcovschi, M. Yiwere, P. Corcoran, and H. Cucu, "Adaptation of Whisper models to child speech recognition," 2023, *arXiv*. doi: 10.48550/ARXIV.2307.13008.

[11] D. P. Kingma and J. Ba, "Adam: A Method for Stochastic Optimization," 2014, doi: 10.48550/ARXIV.1412.6980.

[12] A. Hannun, "Sequence Modeling with CTC," *Distill*, vol. 2, no. 11, p. e8, Nov. 2017, doi: 10.23915/distill.00008.

[13] A. Graves, S. Fernandez, F. Gomez, and J. Schmidhuber, "Connectionist Temporal Classification: Labelling Unsegmented Sequence Data with Recurrent Neural Networks".

[14] A. Conneau, A. Baevski, R. Collobert, A. Mohamed, and M. Auli, "Unsupervised Cross-lingual Representation Learning for Speech Recognition," Dec. 15, 2020, *arXiv*: arXiv:2006.13979. Accessed: Nov. 02, 2023. [Online]. Available: http://arxiv.org/abs/2006.13979

[15] A. Conneau *et al.*, "Unsupervised Cross-lingual Representation Learning at Scale," in *Proceedings of the 58th Annual Meeting of the Association for Computational Linguistics*, Online: Association for Computational Linguistics, 2020, pp. 8440–8451. doi: 10.18653/v1/2020.acl-main.747.

[16] J. Seo, M. Kang, S. Byun, and S. Lee, "ManWav: The First Manchu ASR Model," Jun. 19, 2024, *arXiv*: arXiv:2406.13502. Accessed: Aug. 07, 2024. [Online]. Available: http://arxiv.org/abs/2406.13502

[17] R. He *et al.*, "On the Effectiveness of Adapter-based Tuning for Pretrained Language Model Adaptation," Jun. 06, 2021, *arXiv*: arXiv:2106.03164. doi: 10.48550/arXiv.2106.03164.

[18] K. Nowakowski, M. Ptaszynski, K. Murasaki, and J. Nieuważny, "Adapting Multilingual Speech Representation Model for a New, Underresourced Language through Multilingual Fine-tuning and Continued Pretraining," *Inf. Process. Manag.*, vol. 60, no. 2, p. 103148, Mar. 2023, doi: 10.1016/j.ipm.2022.103148.

[19] Y. Hou, "Fine-tune and deploy a Wav2Vec2 model for speech recognition with Hugging Face and Amazon SageMaker | AWS Machine Learning Blog," aws.amazon.com. Accessed: Jun. 14, 2024. [Online]. Available: https://aws.amazon.com/blogs/machine-learning/fine-tune-and-deploy-a-wav2vec2-

model-for-speech-recognition-with-hugging-face-and-amazon-sagemaker/

[20] "Fine-Tune MMS Adapter Models for low-resource ASR." Accessed: Mar. 18, 2024. [Online]. Available: https://huggingface.co/blog/mms_adapters

[21] "Fine-Tune XLSR-Wav2Vec2 for low-resource ASR with 🤗 Transformers." Accessed: Mar. 18, 2024. [Online]. Available: https://huggingface.co/blog/fine-tune-xlsr-wav2vec2